\documentclass[journal]{IEEEtran}

\usepackage{amsmath,amssymb}
\usepackage{algorithm,algpseudocode}
\usepackage{booktabs,balance}
\usepackage{bm}
\usepackage{color}
\usepackage{courier}
\usepackage{epsfig}
\usepackage{graphicx}
\usepackage{helvet}
\usepackage{multirow}
\usepackage{times}
\usepackage{url}
\usepackage{cite}

\newcommand\ie{\emph{i.e.}}

\newcommand\eg{\emph{e.g.}}

\newcommand{\red}[1]{\textcolor{red}{#1}}
\newcommand{\blue}[1]{\textcolor{blue}{#1}}

\ifCLASSINFOpdf\else\fi

\begin{document}

\title{Edge-Aware Deep Image Deblurring}

\author{Zhichao~Fu, Tianlong~Ma, Yingbin~Zheng, Hao~Ye, Jing~Yang, and Liang~He
\thanks{Z. Fu, T. Ma, J. Yang, and L. He are with East China Normal University, Shanghai, China. (e-mail: 51184506008@stu.ecnu.edu.cn; tlma@cs.ecnu.edu.cn; jyang@cs.ecnu.edu.cn; lhe@cs.ecnu.edu.cn)}
\thanks{Y. Zheng and H. Ye are with Videt Tech Ltd., Shanghai, China. (e-mail: yingbin.zheng@videt.cn; hao.ye@videt.cn)}
}

\maketitle

\begin{abstract}
Image deblurring is a fundamental and challenging low-level vision problem. Previous vision research indicates that edge structure in natural scenes is one of the most important factors to estimate the abilities of human visual perception. In this paper, we resort to human visual demands of sharp edges and propose a two-phase edge-aware deep network to improve deep image deblurring. An edge detection convolutional subnet is designed in the first phase and a residual fully convolutional deblur subnet is then used for generating deblur results. The introduction of the edge-aware network enables our model with the specific capacity of enhancing images with sharp edges. We successfully apply our framework on standard benchmarks and promising results are achieved by our proposed deblur model.
\end{abstract}

\section{Introduction}
\label{sec:intro}

As a branch of image degradation, image blur is a common phenomenon in the realistic shooting scene. In general, blur factors are complex and varied in parts of the image. For example, different choices of aperture size and focal length can lead to Gaussian blur, error operations (such as out of focus), camera shake, and complex scenarios with moving objects that may result in both Gaussian blur and motion blur. It is difficult to confirm the blur reason because of concurrent situations. Besides, blur inversion is a quite ill-posed problem, as a blurry image may correspond to multiple possible clear images. Therefore, the single image blind deblurring is a very challenging low-level vision problem.

Early works for image deblurring depend on various strong hypotheses and natural image priors~\cite{Szeliski2010Computer}. Then some uncertain parameters in the blur model will be certain, such as the type of blur kernel and additive noise \cite{Chan1998Total,Goldstein2012Blur}. However, in the real world scenario and applications, these simplified assumptions on sampling scene and blur model may lead to bad performance. Furthermore, these methods are computationally expensive and usually need to tune a large number of parameters.

In recent years, the applications of deep learning and generative networks on the computer vision tasks have created a significant breakthrough in many research fields. Many regression networks based on Convolutional Neural Networks (CNN) were proposed for image restoration tasks, including a few approaches to handle the image deblurring problem~\cite{Sun2015Learning, Nah2016Deep, Noroozi2017Motion, Tao2018Scale}. Compared to the traditional methods, deep-learning-based approaches have a lower dependence on apriori knowledge, and new models can reconstruct images more accurately both in global and local scales. Early networks are applied to replace the single step of traditional methods, \eg, estimating the kernel or deblurring with a fixed and known kernel \cite{Sun2015Learning,Dong2017From}. More recent works implement an end-to-end learning approach to handle space-variant blur and have achieved state-of-the-art performance~\cite{Nah2016Deep, Noroozi2017Motion, Tao2018Scale}.

There are still some issues of previous deep neural network architecture for image deblurring.
Firstly, although neural networks using deeper architectures are usually efficient, it is hard to interpret the effect of a single component in these networks.
Moreover, the evaluation metrics used in the image restoration tasks, such as peak signal noise ratio (PSNR) and structural similarity index (SSIM), are generally based on pixel-wise or feature-wise differences between the clear natural image and the processed image, tending to enhance the mathematical similarity rather than the human subjective perceptual quality.
PSNR measures image quality by calculating the Mean Squared Error (MSE), which still exists a gap with the assessment of the human visual system. SSIM models human visual quality on several components (such as luminance, contrast, and structure). These components can be used to evaluate visual quality, but still inherently one-sided evaluation on the complexity of human vision.

\begin{figure}[t]
\centering
\includegraphics[width=.99\linewidth]{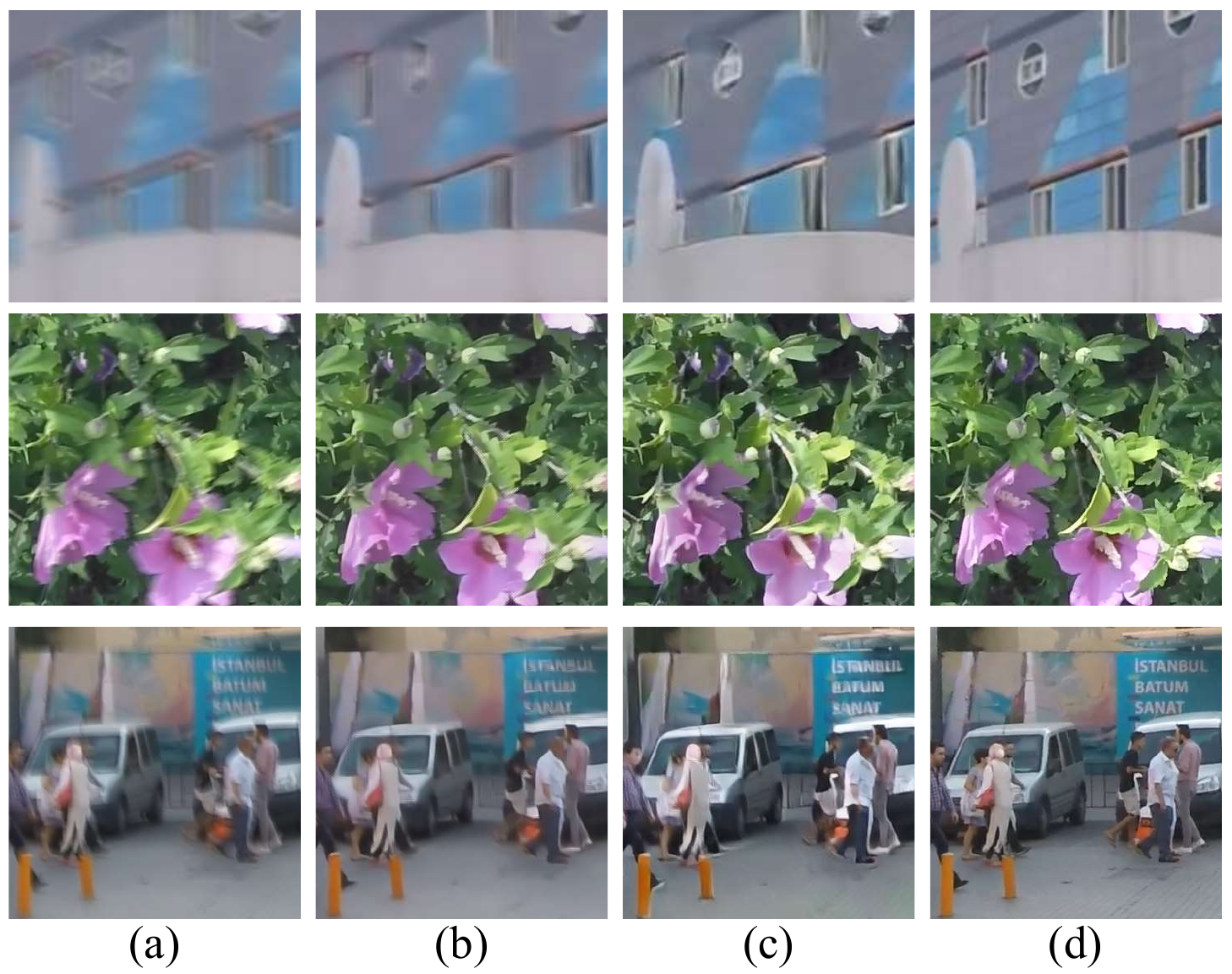}
\caption{(a) blurry images; (b) deblur images without edge information; (c) deblur images with edge; (d) ground truth clear images.}
\label{fig1}
\end{figure}

In this paper, we focus on not only the fitting effect but also the perceptual factors to improve the ability of networks.
The human visual sensitivity to the various frequencies of visual stimuli is measured by the contrast sensitivity functions, which can be an estimation of human visual perception abilities~\cite{Palmer1999Vision}. \cite{Bex2009Contrast} has shown contrast sensitivity functions depend on edge as well as high-frequency structure. Therefore, the reconstruction of edge information in degraded images is the key component to make objects in restored images more recognizable.
As a key component of high frequency, edge information should be incorporated to deal with the deblurring task.
{The edge information has a direct association with visual cognition.}
The clear parts in natural images usually keep their edges sharp and smooth, while blurred regions usually have vignette edges.
{Particularly, The edge information from patches with gaussian blur is weak, as gaussian blur can reduce the edge gradient. And in motion-blurred parts, ghosting images caused by motion break the integrated edge information and make edges unstable and dense. All of these deficient edge components can guide the model to focus on corresponding parts and restore the sharp results. Therefore, edge information is quite related to the expected deblurred results.}
{We are also inspired by the image transformation tasks such as RGB-D semantic segmentation, that incorporating an extra depth channel greatly boosts the performance of the visual system.
Although the depth and edge clues reflect different aspects of the physical structure in an image, we believe it is important to help deblurring as a part of the input.
As blurred image patches are usually quite different on visual modal with clear images and edge maps may contain complementary patterns to original blurry images.
}
{Based on these observations, we combine the edge factor into the image deblurring model.}
The proposed edge-aware deblur network (EADNet) has two phases, \ie, extracting high-frequency edge information and edge-aware deblurring. For each phase, we design a single subnet for the outputs.

The highlight of our work is, our deblurring model separate high-frequency information factors from the end-to-end model, enforcing the network optimized towards specific visual effects, which enhance the interpretability of entire networks.
Although objective optimizes some general metrics like PSNR and SSIM is relatively effective for deblurring, they are not totally adopting human perceptual demand. Our model is trained for sharp edges in deblurred images, which is more helpful for human sensing and recognition.

\begin{figure}[t]
\centering
\includegraphics[width=.95\linewidth]{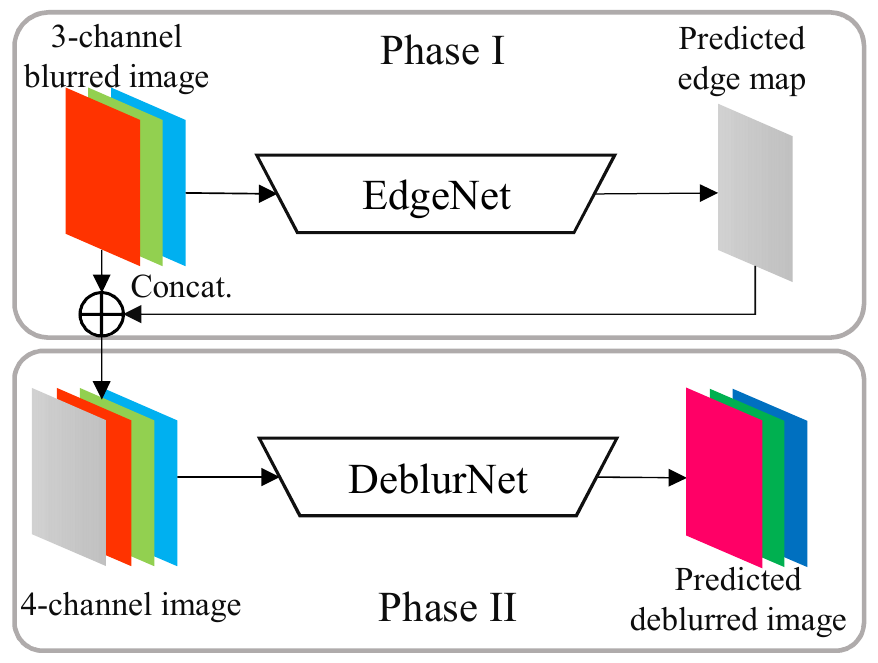}
\caption{The pipeline of the proposed edge-aware deblur network. The two-phase processes are shown from top to bottom. Phase I: EdgeNet to extract the edge map from the original blurry image. Phase II: DeblurNet with the 4-channel input image by concatenating the original blurry image and the edge map and then output the deblur image.}
\label{fig:two-phase deblur layout}
\end{figure}

\section{EADNet}
\label{sec:Framework}
The perceptual quality of restored images is important for evaluating image restoration methods or models. However, it is usually difficult to reach a subjective assessment. First of all, the perceptual quality is defined by human evaluation, while it is a heavy burden for a human to distinguish the high perceptual quality images from the low quality distorted ones in a subjective way. Besides, mainstream objective metrics for deblurring are full-reference, such as peak signal noise ratio (PSNR) and structural similarity index (SSIM), judging restored images by comparing them with original natural ones. Sometimes some learning-based methods may reach high scores on these metrics, but their deblurred results maybe not sharp but just similar to original images. Therefore, more perceptual factors should be considered to improve the substantial capability of the networks. This idea motivates us to build our model and try to enforce model deblurring ability interpretably.

Human recognizes objects by high-frequency components in the images, and the edge is a representation of high-frequency information. The goal of our method is to restore blurred images and make deblurred images with more sharp edges. It is designed to work as a two-phase model. As shown in Figure \ref{fig:two-phase deblur layout}, the EADNet model includes two subnets, namely the \emph{EdgeNet} and \emph{DeblurNet}. The EdgeNet is a network served for Phase I by detecting edges from the blurry image. Then the edge mapping will be concatenated to the original blurry image, as an extra input channel of the next phase. In Phase II, the DeblurNet uses this 4-channel image to deblur the fuzzy part and enforce the edges with the aid of the edge mapping, and finally outputs the deblurred images.
We will start by introducing the details of both networks in the section, followed by their training strategy in Section \ref{sec:training}.

\begin{figure}[t]
\centering
\includegraphics[width=0.98\linewidth]{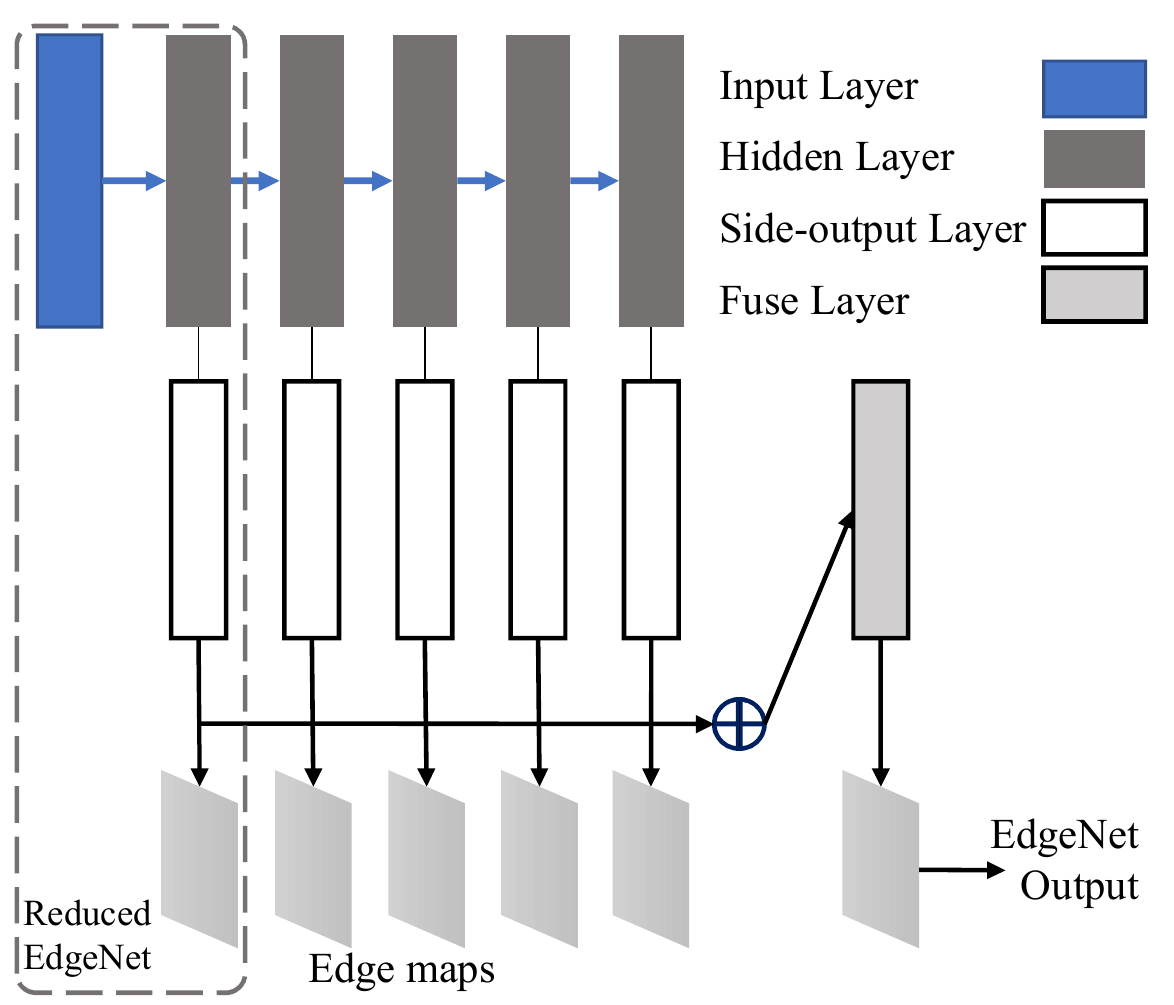}
\caption{ The network architecture of EdgeNet.}
\label{fig:EdgeNet}
\end{figure}

\begin{figure*}[t]
\centering
\includegraphics[width=.75\linewidth]{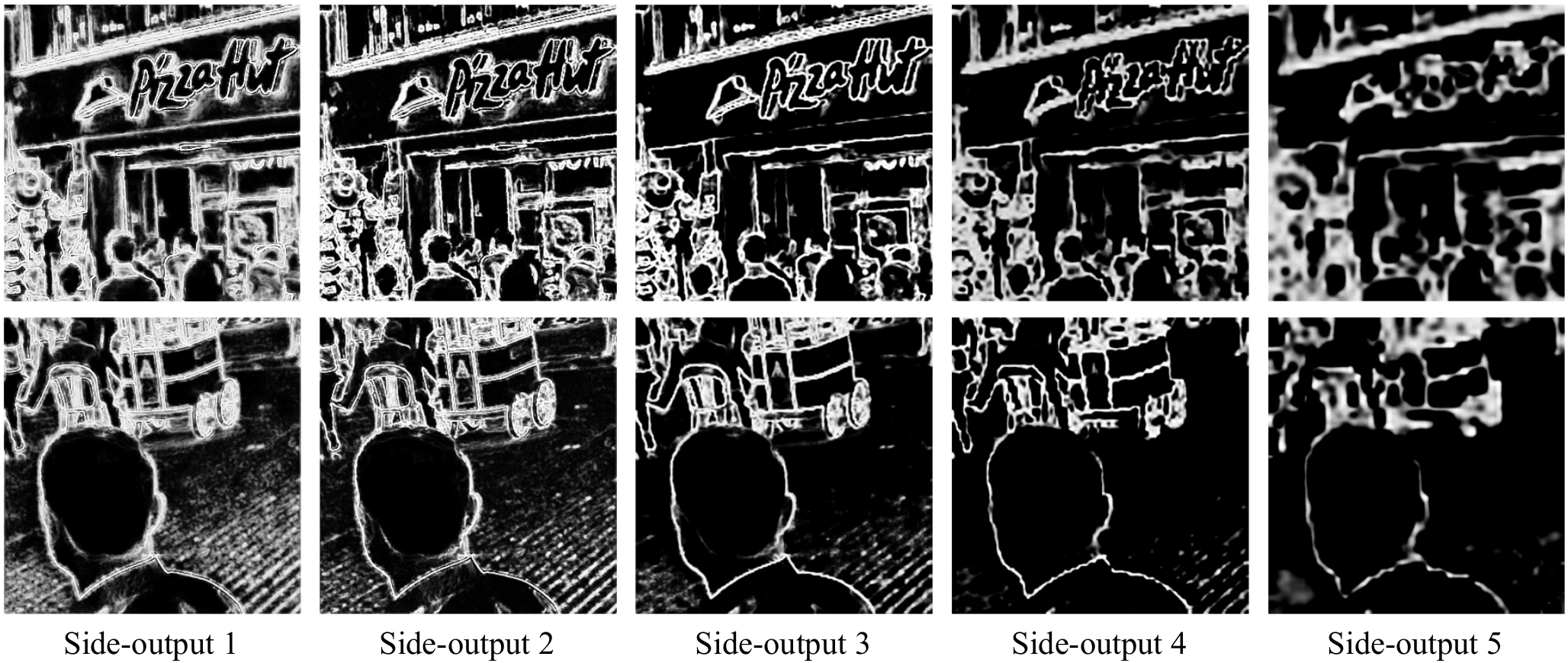}
\caption{{Responses on side-output layers. From shallow to deep, the side outputs become progressively coarser.}}
\label{fig:example-sideoutput}
\end{figure*}

\subsection{EdgeNet}

For the edge detection network in the deblurring pipeline, initially, we would like to employ a function or simple module rather than a neural network, such that we can enforce the network capacity without massive extra computation. However, we found that traditional methods like Canny detector~\cite{Canny1986A} limited by some artificial constraint or threshold and do not have proper adaptation.
Then we switch to the network-based models and a trimmed VGG-based network~\cite{simonyan2014deep}, \ie, the Holistically-nested Edge Detection (HED, \cite{Xie2015Holistically}) is chosen as our basic structure.
As shown in Fig. \ref{fig:EdgeNet}, the side-output layers are inserted after convolutional layers, serving multi-scale and multi-level outputs. At last, one additional weighted-fusion layer will combine outputs from multiple scales.

Note that even the original edge detection network is with very high detection performance, its applications on the blurry images are not as good as the clear images.
We observe that these multi-scale side-output layers from EdgeNet have an interesting characteristic: the first side-output layers preserve the detailed and local edges as the last side-output layers (see Fig. \ref{fig:example-sideoutput}). Inspired by this, we use different strategies during the training and processing stage. The whole network is used for training, so as to acquire multi-scale edge detection capacity. While in the testing stage, we use a reduced subnet, which only remains the input layer to the side-output layer 1. This choice makes a trade-off between performance and model efficiency, which keeps enough edge information and requires less computation resources during inference.

\subsection{DeblurNet}
\begin{figure}[t]
\centering
\includegraphics[width=\linewidth]{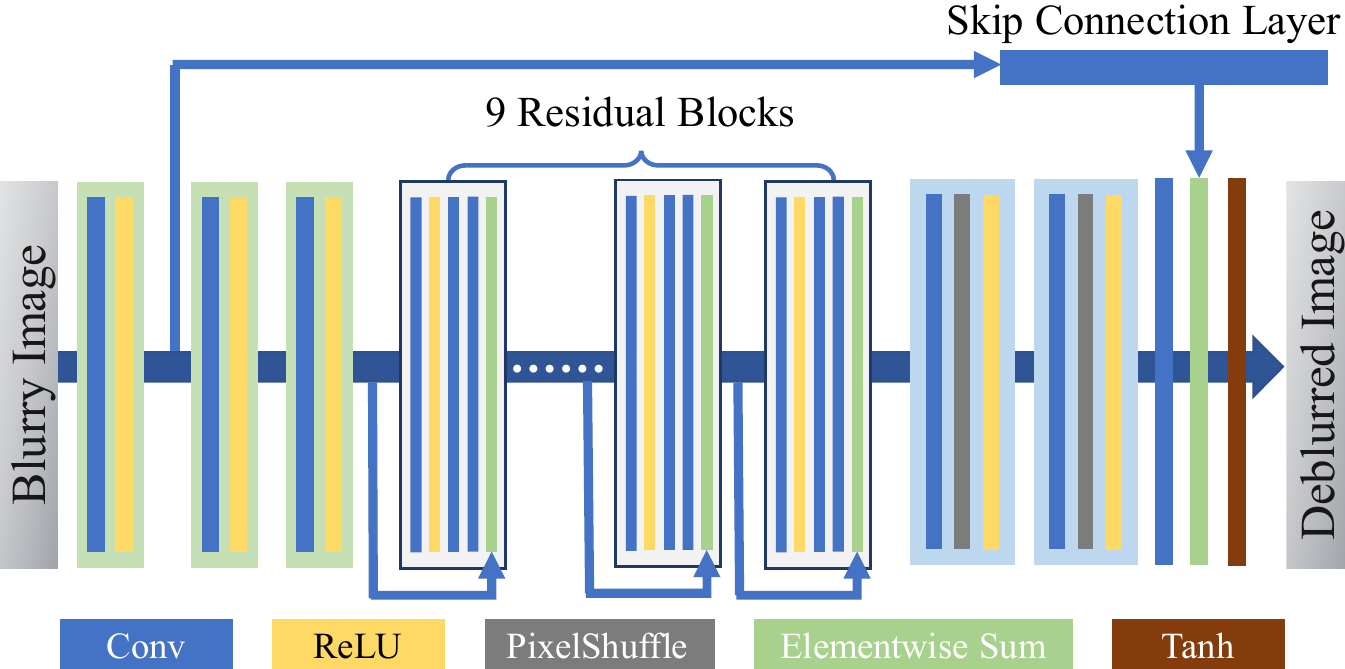}
\\(a)\\
\includegraphics[width=.95\linewidth]{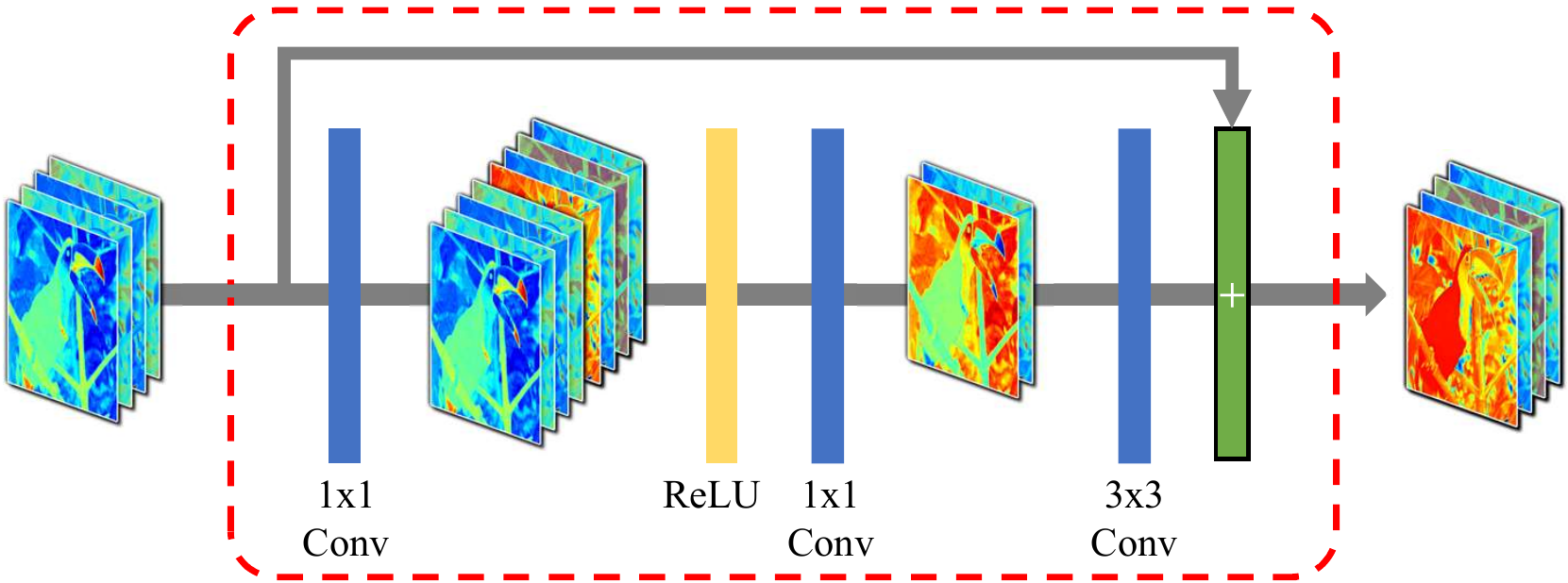}
\\(b)
\caption{(a) The network architecture of DeblurNet; (b) The Residual Block architecture employed in DeblurNet.}
\label{fig:deblurnet architecture}
\end{figure}

As shown in Figure \ref{fig:deblurnet architecture}(a), a generative CNN architecture is employed as the DeblurNet. The network consists of three convolution blocks, nine residual blocks, two upsampling convolution blocks, and a global skip-connection convolutional layer. Besides, weight normalization is applied to convolutional layers for easier training.

In the first convolution block, we use a big convolution kernel with kernel size 9 and stride 4 to extract low-level feature mappings the same width and height as original images. Then two convolution blocks work as downsampling blocks, generating half-size feature mappings using small kernel (kernel size 3, stride 2). All these convolution blocks use ReLU for activation after convolutional layers.

{The residual-based network is a common structure for low-level vision tasks.
As blur degradation is usually with complicated blur kernels, it is necessary to use networks with strong representation capacity. Therefore, we choose the residual block introduced in~\cite{yu2018wide} with wider activation and low-rank convolution (Fig. \ref{fig:deblurnet architecture}(b)) and set a residual block number to 9. In each residual block, channel numbers are expanded by the first convolutional layer using $1 \times 1$ kernel and then apply ReLU activation. After that, we employed a group of efficient linear low-rank convolution, a stack of one $1 \times 1$ convolution reducing channel numbers and one $3 \times 3$ convolution performing spatial-wise feature extraction.}

{We also designed the upsample convolutional blocks by using sub-pixel convolution, including pixelshuffle operation and a convolutional layer. These upsampling blocks enlarge 2-D feature mappings shape (width and height) and compress the channel numbers.}

Finally, a global skip-connection structure generates the final output. Two convolutional layers using $9 \times 9$ kernels take in low-level features from the first convolution block and high-level features from residual body respectively. Then we apply element-wise summation on two 3-channel outputs before the final Tanh activation layer. {The architecture designs keep our DeblurNet using fewer parameters and memory but performing stronger representation ability.}

\begin{figure*}[t]
\centering
\includegraphics[width=\linewidth]{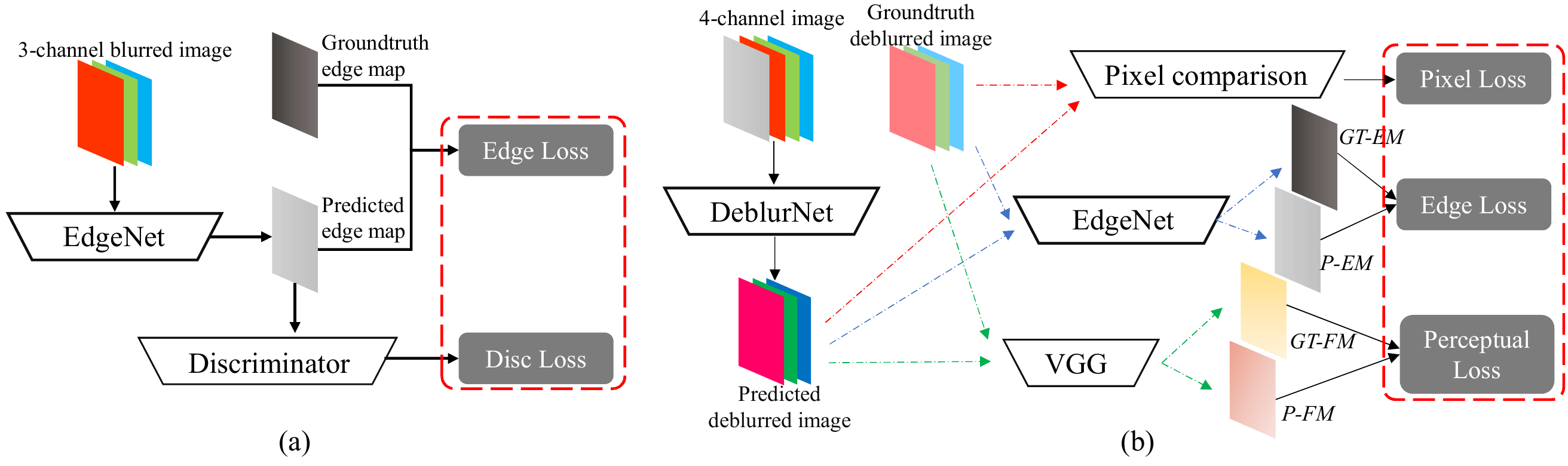}
\caption{The training process for the parameters of EdgeNet (a) and DeblurNet (b). \emph{GT-EM} and \emph{P-EM} indicate the ground truth and predicted edge map. \emph{GT-FM} and \emph{P-FM} indicate the ground truth and predicted feature map.}
\label{fig:network training}
\end{figure*}

\section{Network Training}
\label{sec:training}

\subsection{Blurry Images and Edge Map Generation}
\label{subsec:blurry edge generation}
Generally speaking, when training an edge detection network, clear images are employed as input data pairs and edge maps as the ground truth.
{We can observe that different datasets such as the GOPRO and Kohler images are quite different on the blind kernel modal. If we only use one of them for training, the results should be biased. It is hard to build a robust CNN model with the kernels that have not appeared in the training set.
Moreover, our EdgeNet is designed to extract edges from blurry images and it is hard to find an off-the-shelf dataset offering blurry images with clear edge maps.
Inspired by the data augmentation and image synthesis methods used in many other vision tasks, we build the training dataset by generating from clear images in MS COCO dataset~\cite{lin2014microsoft} and use them to train both the EdgeNet and DeblurNet.
}

\begin{figure}[t]
\centering
\includegraphics[width=.99\linewidth]{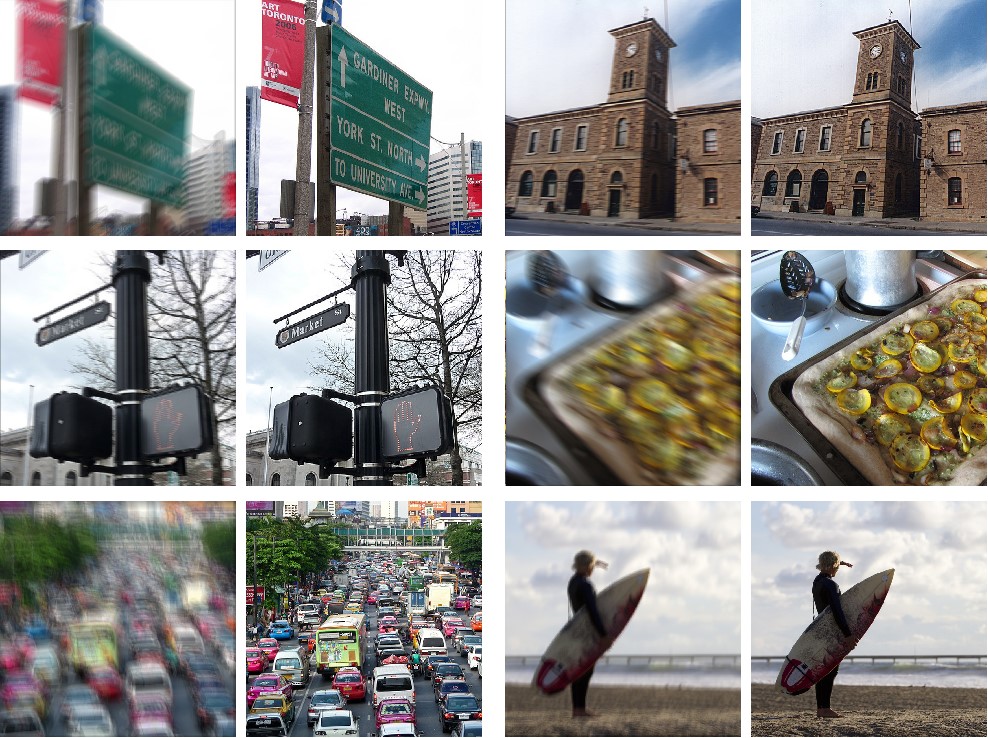}
\caption{{Some synthesize blurry images (left in each group) by random generated blur kernels on clear MS COCO images (right).}}
\label{fig:synthesize}
\end{figure}

In order to keep the adaptability and robustness of the network in the complex image scenario, we randomly add the Gaussian blur or motion blur to generate the blurry input images. Specifically, for the motion blur, we choose the method proposed by \cite{boracchi2012modeling}, which firstly generates a random trajectory vector by the Markov process and then applies sub-pixel interpolation to trajectory vectors to generate the blur kernel.
{As illustrated in Fig. \ref{fig:synthesize}, the blur modal of many synthetic images with randomly generated blur kernels are close to images that are real acquisitions.
With the help of this synthetic dataset, our model can learn the capacity to deal with various kinds of blur images.}

To generate the ground truth edge map, we choose a classic method, \ie, the Canny edge detector \cite{Canny1986A}. Comparing to the annotation by the human, the Canny edges are dependent on strong artificial thresholds, usually not directly connected, and exhibiting spatial shift and inconsistency. However, the Canny detector is less time consuming, and experiments validate that the Canny edges are still useful for training the EdgeNet.

\subsection{Phase I: EdgeNet Training}
\label{subsec:edgenet training}
Compared with the original HED model, we make some changes during the training process. The modified training framework is shown in Figure \ref{fig:network training}(a). We introduce a discriminator network to build adversarial training. As discriminator in generative adversarial networks always has strong image distribution fitting abilities, it can also help EdgeNet to learn how to extract blurred edges from blurry images. The architecture of this discriminator is similar to PatchGAN~\cite{Isola2016Image}, and all the convolutional blocks have a convolutional layer followed by a Spectral Normalization layer~\cite{Miyato2018Spectral} and LeakyReLU~\cite{xu2015empirical}.
We formulate the loss function as follows, including edge loss term $\mathcal{L}_{Edge}$ and adversarial loss $\mathcal{L}_{Disc}$ with trade-off $\lambda$:
$$\mathcal{L}=\mathcal{L}_{Edge}+\lambda\cdot\mathcal{L}_{Disc}$$
The edge loss is calculated based on class-balanced cross-entropy loss $l_{CBCE}$ mentioned in \cite{Xie2015Holistically},
$$\mathcal{L}_{Edge}=
    \sum\limits_{i}^{n}l_{CBCE}(\phi_{side_i}(I_B), E)
    +l_{CBCE}(\phi_{fuse}(I_B), E),$$
where $E$ and $I_B$ are ground truth edge mapping and input blurry image, $\phi_{side_i(I_B)}$ and $\phi_{fuse}(I_B)$ are output edge mappings from side output layer $i$ and final fuse layer respectively.

The adversarial loss term $\mathcal{L}_{Disc}$ is calculated as vanilla GAN~\cite{goodfellow2014generative}, where $D_{\theta_D}$ and $G_{\theta_E}$ represent the network parameters for discriminator and generator (\ie, EdgeNet).
$$\mathcal{L}_{Disc}=-\log(D_{\theta_D}(G_{\theta_E}(I_B)))$$

In our implementation, we use the discriminator network as a training accelerator. At the beginning of EdgeNet training, we set a quite small $\lambda$ to 0.05, avoiding the overfitting to training data. After 50 epochs, $\lambda$ is set to 0, which means the discriminator only used in pretraining processing.

\begin{figure*}[t]
\centering
\includegraphics[width=.99\linewidth]{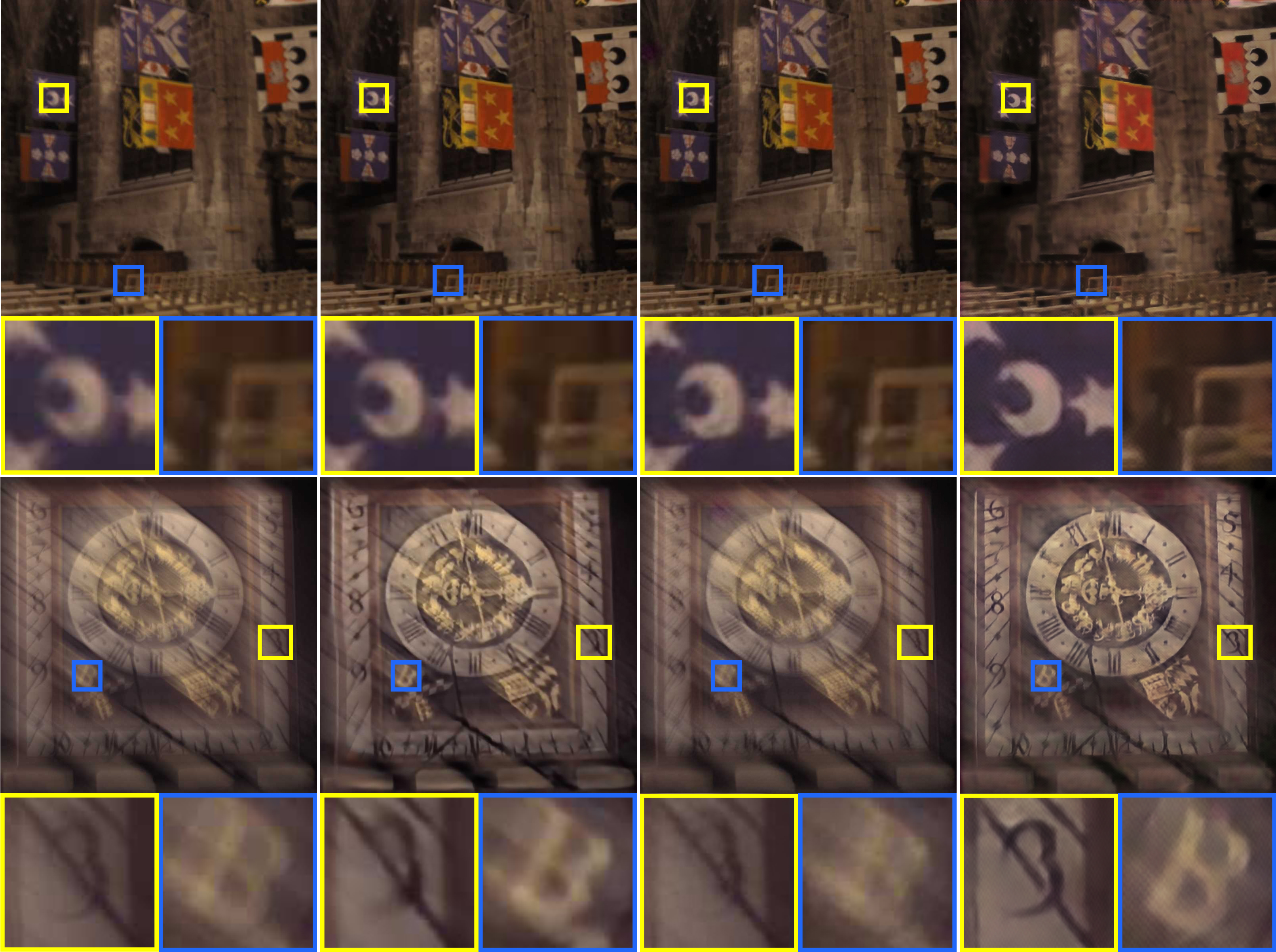}
\caption{{Comparison of visual details on Kohler test dataset. From left to right: Input, SRN\cite{Tao2018Scale}, DeblurGAN v2(Inception)~\cite{kupyn2019deblurgan}, and the proposed EADNet.}}
\label{fig:visual kohler}
\end{figure*}

\begin{figure*}[t]
\centering
\includegraphics[width=.99\linewidth]{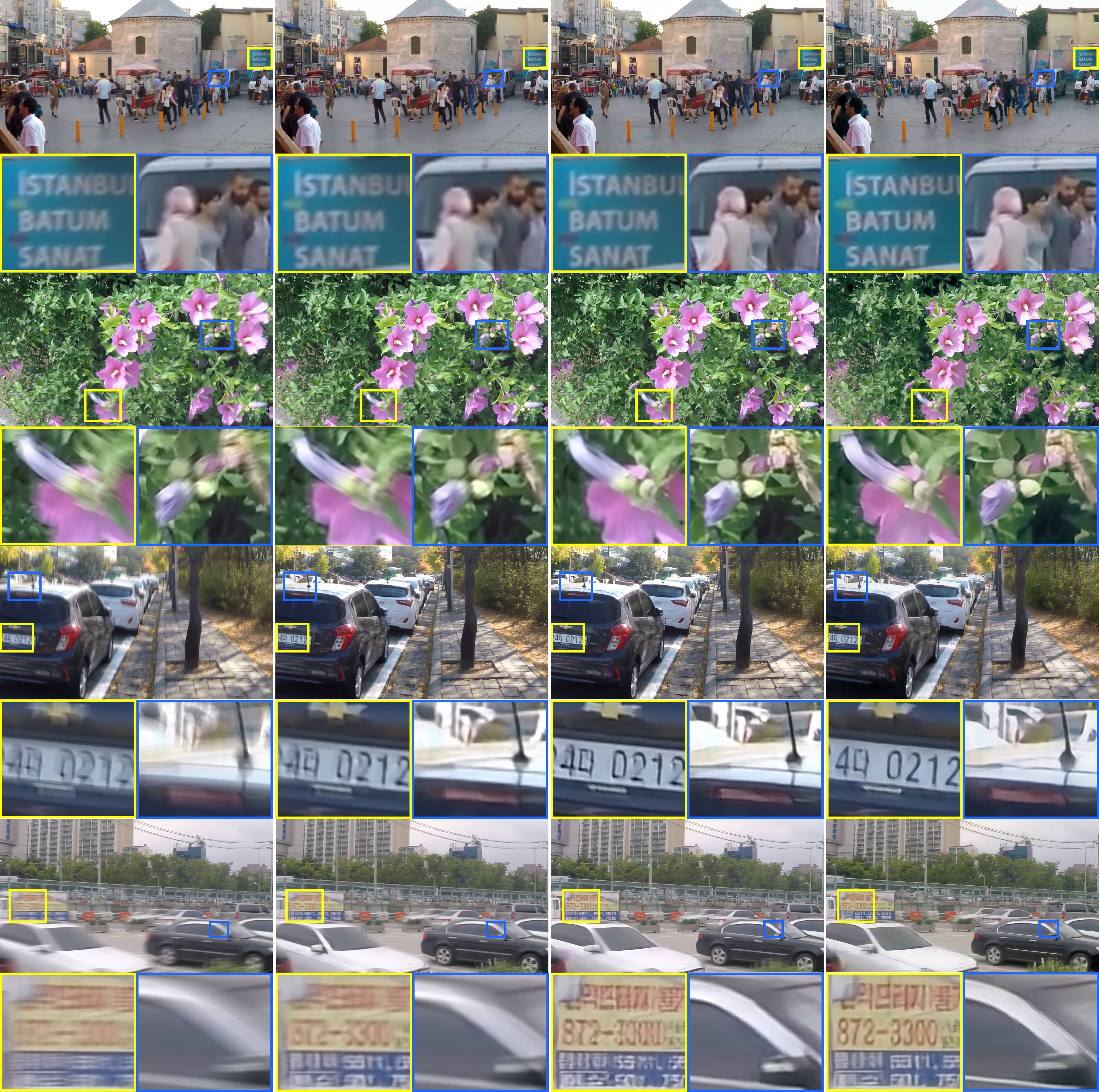}
\caption{{Comparison of visual details on GoPro test dataset. From left to right: Input, SRN\cite{Tao2018Scale}, DeblurGAN v2(Inception)~\cite{kupyn2019deblurgan}, and the proposed EADNet.}}
\label{fig:visual GoPro}
\end{figure*}

\subsection{Phase II: Edge-Aware Training of DeblurNet}

The training of DeblurNet is directly connected to that of EdgeNet. Not only the edge map channel of input comes from the EdgeNet, but also the EdgeNet will guide the DeblurNet to learn to build sharp edges for deblurred images. As illustrated in Figure \ref{fig:network training}(b), we have three terms in the loss function.
$$\mathcal{L}=
        \lambda_1\cdot\mathcal{L}_{Pixel}
        +\lambda_2\cdot\mathcal{L}_{Perceptual}
        +\lambda_3\cdot\mathcal{L}_{Edge}
$$

The first term is the pixel loss, which is the MSE by pixel-wise comparison. The perceptual loss, inspired by \cite{johnson2016perceptual}, is based on the difference of feature maps from an ImageNet~\cite{ILSVRC15} pre-trained VGG19 network \cite{simonyan2014deep} between the generated and target image. Formally, the perceptual loss is as follows,
$$\mathcal{L}_{Perceptual}=
    \frac{1}{W_jH_j}
    \sum\limits_x^{W_j}
    \sum\limits_y^{H_j}
    (\phi_j(I_C)_{x,y}-\phi_j(G_{\theta_D}(I_B))_{x,y})^2
$$
where the $\phi_j$ represents the $j$-th convolutional layer.
Both the pixel and perceptual terms are considered as content loss and here we use $L_2$ metric to compute the MSE.

The third term is the edge loss, which is similar to the one for EdgeNet training and also based on the class-balanced cross-entropy loss. The edge maps for calculating loss are extracted from blurry inputs $(I_B)$ and DeblurNet deblurred outputs $(I_C)$ respectively. The loss function is defined as follows,
$$\mathcal{L}_{Edge}=
    \sum\limits_{i}^{n}\alpha_i\cdot l_{CBCE}(\phi_{side_i}(I_B), \phi_{side_i}(I_C)),
$$
where $\alpha_i$ is the weight for side-output layer $i$. In our experiment, we set $\alpha_1=0.7$ and $\alpha_i=0.1 (i>1)$ as the tailer version of HED is used.

\section{Experiments}
\label{sec:exp}

\subsection{Datasets}

We evaluate the framework of edge-aware deblur network on two image deblurring benchmarks: GOPRO~\cite{Nah2017Deep} and Kohler~\cite{K2012Recording}.
GOPRO dataset~\cite{Nah2017Deep} consists of 3214 pairs of blurred and sharp images in 720p quality, taken from various scenes (2103 pairs for training and 1111 for testing). The blurry images are generated from clear video images. Kohler dataset~\cite{K2012Recording} is also a standard benchmark dataset for the evaluation of blind deblurring algorithms. The dataset includes 4 clear images, and each of them is blurred with 12 different blur kernels generated with on real camera motion records and analysis. It is played back on a robot platform such that a sequence of sharp images is recorded sampling the 6D camera motion trajectory.

To keep the generic deblurring capacity for our model, we use a mixed dataset during the training process. The final representation of mixed dataset has three parts, \ie, clear images, blurry images, and edge images from clear images. And the mixed dataset has two sources, \ie, MS COCO dataset~\cite{lin2014microsoft} and GoPro training set~\cite{Nah2016Deep}. MS COCO dataset only consists of clear sharp images, so we randomly choose 2000 images, using Canny edge detector to extract edge images from clear images and the method mentioned in Section \ref{subsec:blurry edge generation} to generate blurry images. The GoPro training set has 2103 pairs of clear and blurry images, so we only use Canny to obtain clear edge images.

\begin{table*}[t]
\caption{{Metrics results and parameter number comparison with state-of-the-art approaches. Red color text denotes the top performer and blue denotes the runner-up. The EADNet is with the default setting of the reduced EdgeNet.}}
\label{tab:newscores}
\centering{
\begin{tabular}{l|c|c|c|c|c|c}
\hline
\multirow{2}*{Method} & \multicolumn{2}{|c|}{GOPRO} & \multicolumn{2}{|c|}{Kohler} & \multirow{2}*{Time} & \multirow{2}*{Params}\\
\cline{2-5}
& PSNR & SSIM & PSNR & MSSIM & & \\
\hline
Kim \emph{et al.}~\cite{Kim2013Dynamic} & 23.64 & 0.8239 & 24.68 & 0.7937 & 1 hr & -\\
Sun \emph{et al.}~\cite{Sun2015Learning} & 24.64 & 0.8429 & 25.22 & 0.7735 & 20 min & -\\
Nah \emph{et al.}~\cite{Nah2016Deep} & 29.08 & 0.9135 & 26.48 & 0.8079 & 2.51 s & -\\
SRN~\cite{Tao2018Scale} & 30.10 & \blue{0.932} & \blue{26.80} & \red{0.8375} & 0.67 s & 10.25M\\
\hline
DeblurGAN~\cite{Kupyn2017DeblurGAN} & 28.70 & \red{0.958} & 26.10 & 0.816 & 0.85 s & \blue{6.07M}\\
DeblurGAN v2 (Inception)~\cite{kupyn2019deblurgan} & 29.55 & 0.934 & 26.72 & \blue{0.836} & 0.35 s & 60.93M\\
DeblurGAN v2 (MobileNet)~\cite{kupyn2019deblurgan} & 28.17 & 0.925 & 26.36 & 0.820 & \red{0.04 s} & \red{3.31M}\\
\hline
EADNet (without EdgeNet) & 29.53 & 0.9014 & 25.97 & 0.8189 & \blue{0.16 s} & 8.95M\\
EADNet (with full EdgeNet) & \blue{30.78} & 0.9137 & 26.61 & 0.8297 & 0.23 s & 23.67M\\
EADNet  & \red{31.02} & 0.9123 & \red{26.91} & {0.8325} & 0.18 s & 8.99M\\
\hline
\end{tabular}
}
\end{table*}

\subsection{Model Training}

We use Adam~\cite{Kingma2014Adam} for our model training on both subnets with parameters $\beta_1=0.9$ and $\beta_2=0.999$. The initial learning rate is 0.0005 and decay to one-tenth every 20 epochs. Limited by the memory, we sample a batch of 4 blurry images and crop 256$\times$256 patches randomly for training inputs. In order to save the training time, a co-training method of EdgeNet and DeblurNet is applied to our two-phase deblur model. We first train the EdgeNet with 50 epochs as mentioned in Section \ref{subsec:edgenet training} and then train the DeblurNet.
We implement our model using the PyTorch deep learning library~\cite{Pytorch.org}. The experiments are conducted with Intel Xeon E5 CPU and NVIDIA Titan X GPU.

\subsection{Comparison with State-of-the-arts}

We first compare the EADNet with some previous work or recent state-of-art image deblurring approaches on standard metrics like PSNR, SSIM, and running time. The experiments are running in the same environment and the results are shown in Table \ref{tab:newscores}.

We compare EADNet with two groups of baseline methods. The first group is the state-of-the-art convolution or recurrent network based deblur methods. We can observe that our model outperforms or achieves comparable results with these methods in the different metric scores, and our approach is much faster.
Comparing with the GAN-based methods, the PSNR of EADNet is higher than DeblurGAN~\cite{Kupyn2017DeblurGAN} and DeblurGAN v2~\cite{kupyn2019deblurgan}, while the SSIM of EADNet is not as good as DeblurGAN v2 (Inception).
As mentioned earlier, full-reference metrics for image restoration tasks like PSNR and SSIM may not be perfect. Therefore, we also evaluate the visual effects of deblurred images and some visual details comparison are shown in Fig. \ref{fig:visual kohler} and \ref{fig:visual GoPro}. Here we choose two of the recent
works with high metric, \ie, SRN~\cite{Tao2018Scale} and DeblurGAN v2 (Inception)~\cite{kupyn2019deblurgan} for comparison. From these figures, we can find this network eliminates most blurry structure, and deblurred images from our models are with more sharp and smooth edges.
Our model is able to handle Gaussian blur and motion blur at the same time so that the edges in our deblurred images are sharper. The DeblurNet in our model has indeed learned the specific capacity to generate more sharp images than the compared methods.

As shown in Table \ref{tab:newscores}, the parameter number of EADNet with different settings is between that of the baseline DeblurGAN v2 with MobileNet and Inception~\cite{kupyn2019deblurgan}, and in the same order of magnitude with SRN~\cite{Tao2018Scale}. Within the architecture of our networks, the reduced EdgeNet has an even negligible parameter number (0.04M vs 14.72M) but is more efficient than the full EdgeNet, and the DeblurNet consumes less than 9M parameters.

\begin{figure}[t]
\centering
\includegraphics[width=.99\linewidth]{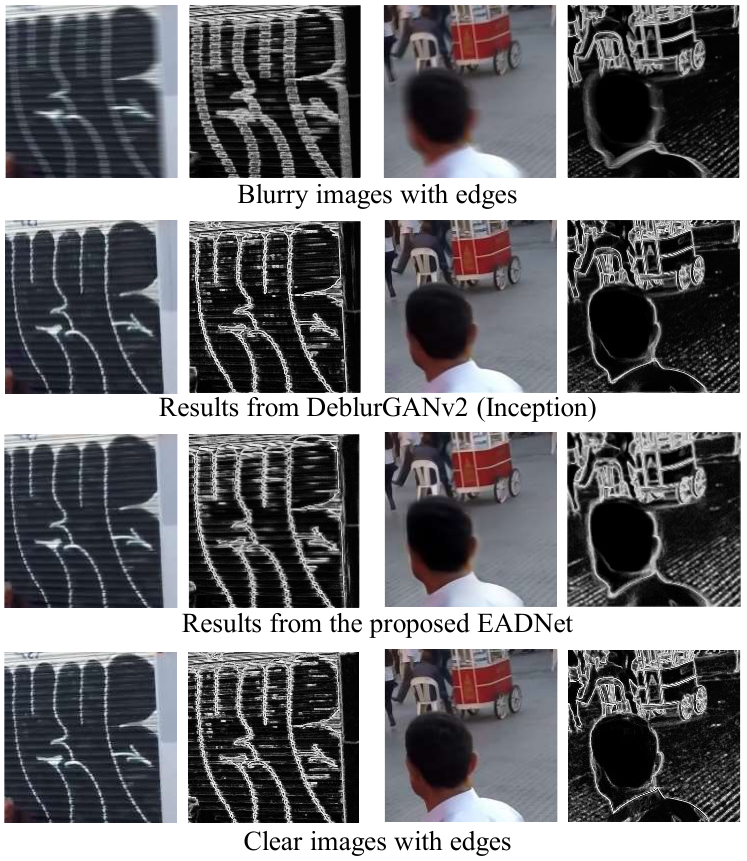}
\caption{The patches cropped from images (left in each group) and their edge details (right).}
\label{fig:patch comparison}
\end{figure}

\subsection{Ablation Study}

In this section, we focus on evaluating the impact of the edge information.
We first compare the edge maps obtained by the images and some patch-level results are shown in Fig. \ref{fig:patch comparison}. The pixels' brightness in deblurred edge maps is higher than those in blurry edge maps and deblurred edge maps have more clear lines. And the edge maps from deblurred results are similar to these from clear images. From the figure, we also see that the edge images from EADNet are more similar to the clear images' edge than the ones from DeblurGAN v2 (Inception)~\cite{kupyn2019deblurgan}. This appearance reveals that our network can generate deblurred images with sharp and clear edges.

\begin{table}[t]
\caption{Results on different EdgeNet settings of the EDANet model.}
\label{tab:experiment ablation}
\centering{
\begin{tabular}{c|c|c|c|c}
\hline
\multirow{2}*{Method} & \multicolumn{2}{|c|}{GOPRO} & \multicolumn{2}{|c}{Kohler} \\
\cline{2-5}
    & PSNR & SSIM & PSNR & MSSIM\\
\hline
EADNet without EdgeNet & 29.53 & 0.9014 & 25.97 & 0.8189 \\
\hline
EADNet with EdgeNet (Layer 1) & \bf{31.02} & 0.9123 & \bf{26.91} & \bf{0.8325} \\
EADNet with EdgeNet (Layer 3) & 30.75 & 0.9096 & 26.70 & 0.8288 \\
EADNet with EdgeNet (Layer 5) & 30.26 & 0.9050 & 26.36 & 0.8239 \\
EADNet with EdgeNet (Full) & 30.78 & \bf{0.9137} & 26.61 & 0.8297 \\
\hline
\end{tabular}
}
\end{table}

We also conduct a set of experiments without the edge information (including the input edge mapping and edge loss) and the results are illustrated in the first row of Table \ref{tab:experiment ablation}. {We also test several settings with the edge, \ie, by using the full EdgeNet and using the reduced versions from different side-output layers. Using only the edge maps from side-output layer 1 (\ie, Reduced EdgeNet, as shown in the left part of Fig. \ref{fig:EdgeNet}), we can already get higher or comparable PSNR/SSIM, while the computational costs are greatly reduced.} And the substantial performance gains over the results without edge and the deblur results (as illustrated in Fig. \ref{fig1}) confirm the effectiveness of using EdgeNet as the basic elements for deep image deblurring tasks.

\begin{figure}[t]
\centering
\includegraphics[width=.7\linewidth]{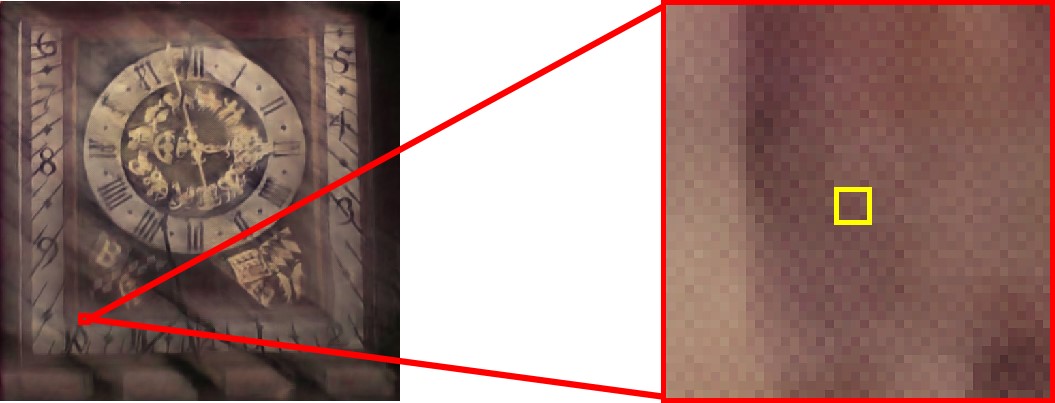}
\caption{Perceptual texture patch from the sample deblurred results.}
\label{fig:artifacts}
\end{figure}

{We observe that both in Fig. 7 and 8 can find dense tiny artifact units, which make enlarged images have paint canvas texture. As for the reason, we think it is mainly caused by perceptual loss used in our training stage. We find similar artifacts in many previous image restoration works with perceptual loss \cite{johnson2016perceptual,sajjadi2017enhancenet,wang2018perceptual}. When perceptual loss is used on the image restoration task, the global texture styles including stripes, shapes, and colors are learned from the training set and transferred into the testing images. All of these textures are also repeated and comprised of obvious units. Perceptual loss is useful for image contrast and better global visual quality, but cause these artifacts from the repeat textures.
Our designs on DeblurNet aim to control the artifacts.
When building the perceptual loss term, we use the early layers to compute perceptual loss and produce deblur images that are visually indistinguishable from the input image.
We notice that previous works such as \cite{johnson2016perceptual} also employed a similar strategy.
The edge loss also helps to reduce the artifacts.
An example of artifact details for our method is given in Fig. \ref{fig:artifacts}, where we zoom the deblurred image patch in the red box and them increase the contrast of the patch for better observation. We can find that artifacts in the whole image are not obvious from global vision.}

\section{Conclusions}
\label{sec:conclusion}
We explored the real demand for human vision in the deblurring task and validated that edge matters in the deep image deblurring system. A two-phase edge-aware deblur network composed of an edge detection subnet as well as a deblur subnet is proposed.
One important goal in this work is to build a novel deblurring model with the capacity of making images
with sharp edges;
the deblurring results from our model have sharp edges, which make objects in images easy to recognize.
We conduct experiments using the EADNet framework on a few benchmark images and demonstrate its superiority in terms of effectiveness and efficiency over previous approaches.

\ifCLASSOPTIONcaptionsoff
  \newpage
\fi

\balance
\bibliographystyle{IEEEtran}
\bibliography{total}

\end{document}